\documentclass[runningheads]{llncs}

\usepackage{eccvabbrv}

\usepackage{graphicx}
\usepackage{booktabs}
\usepackage{multirow}
\usepackage{xcolor}
\usepackage{xspace}
\usepackage{soul}
\usepackage{comment}
\usepackage{subcaption}
\usepackage[accsupp]{axessibility}  

\usepackage{hyperref}

\usepackage{orcidlink}

\begin{document}

\newcommand{\methodName}{\textsc{DiffUE}\xspace}
\newcommand{\vect}[1]{\ensuremath{\mathbf{#1}}}

\title{\textsc{DiffUE}: Enhancing Utility-Unlearnability Trade-off of Unlearnable Examples via Diffusion Autoencoders
}
\titlerunning{\methodName: Enhancing Utility-Unlearnability Trade-off}
\newcommand{\equalcontrib}{\textsuperscript{†}}
\author{
Syed Irfan Ali Meerza$^{1,4,\ast,\dagger}$\orcidlink{0000-0002-3239-1080}
\and
Oktay Ozturk$^{1,\dagger}$\orcidlink{0000-0001-5260-7312}
\and
Amir Sadovnik$^{2}$\orcidlink{0000-0001-9011-7365}
\and
Jian Liu$^{1,3}$\orcidlink{0000-0002-8331-0834}
}

\authorrunning{Meerza et al.}

\institute{University of Tennessee, Knoxville, TN, USA\\ \and
Oak Ridge National University, Knoxville, TN, USA\\ \and
University of Georgia, Athens, GA, USA\\ \and
Virginia Commonwealth University, Richmond, VA, USA
}

\maketitle
\begingroup
\renewcommand{\thefootnote}{\fnsymbol{footnote}}
\footnotetext[1]{Work done while at the University of Tennessee, Knoxville.}
\endgroup

\addtocounter{footnote}{-1}
\renewcommand{\thefootnote}{$\dagger$}
\footnotetext{These authors contributed equally.}
\renewcommand{\thefootnote}{\arabic{footnote}}

\begin{abstract}
AI models are increasingly trained on personal images scraped from social media and public platforms, often without consent, leading to serious privacy violations, such as unauthorized facial recognition and targeted advertising. To counter this, researchers have developed unlearnable examples (UEs), images modified with imperceptible noise to prevent AI models from extracting meaningful information. However, existing UE methods primarily rely on pixel-space noise, which can be bypassed by relearning strategies such as adversarial training, image transformation, and compression. While some techniques improve robustness, they often come at the expense of significant degradation in image utility and perceptual quality. In this paper, we introduce \methodName to overcome these limitations by injecting noise into the semantic space of images instead of the pixel space. Instead of corrupting pixel values, \methodName modifies high-level semantic features of images, ensuring robust unlearnability while preserving visual quality and utility. By leveraging a diffusion-based autoencoder framework to manipulate semantic features, \methodName generates purposeful, natural-looking modifications that effectively resist advanced relearning strategies. Extensive experiments on four datasets, CIFAR-10, CIFAR-100, CelebA-HQ, and ImageNet, as well as a subjective user study, demonstrate that \methodName significantly enhances the trade-off between image quality and unlearnability, offering a more robust and effective solution for safeguarding personal data in an increasingly exploitative AI landscape. 

\end{abstract}

\begin{figure}[t!]
    \centering
    \includegraphics[width=0.9\linewidth]{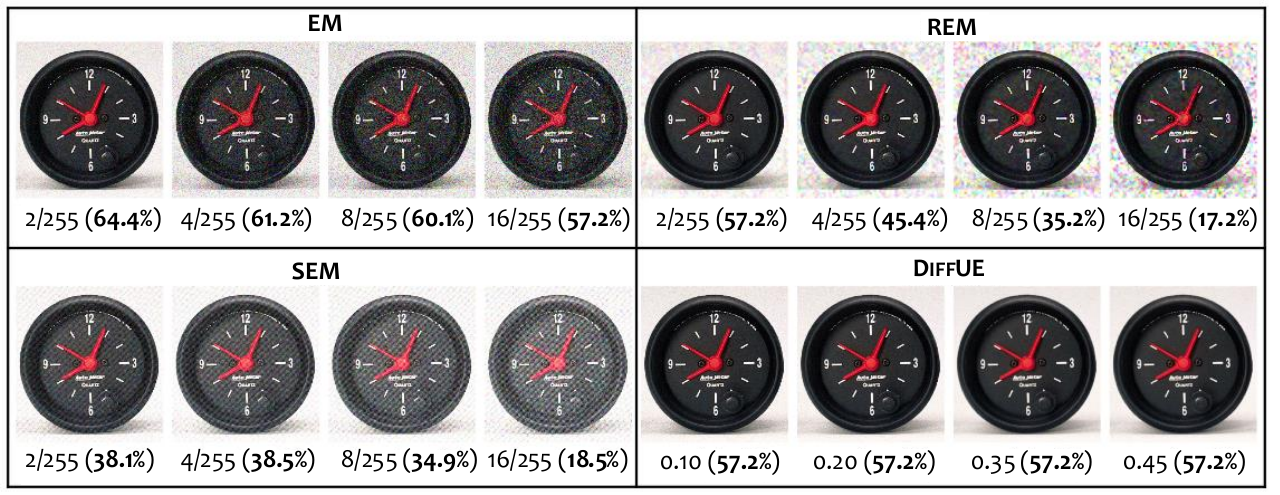}
    \vspace{-2mm}
    \caption{Usability vs. unlearnability comparison under adversarial training with an adversarial perturbation radius of $\rho_a = 4/255$. We compared~\methodName with three baselines: EM~\cite{huang2021unlearnable}, REM~\cite{fu2022robust} and SEM~\cite{liu2024stable}. The number below each image represents the defensive noise radii ($\rho_u$) applied (the defensive noise of \methodName is in the semantic space), with the value in parentheses indicating the test accuracy achieved. Lower accuracy reflects greater unlearnability. Note that we intentionally selected images with a light background to better visualize perturbation artifacts in existing UE methods.}
    \vspace{-6mm}
    \label{fig: utility_vs_usability}
\end{figure}

\vspace{-9mm}
\section{Introduction}
\label{sec:intro}
\vspace{-2mm}

The success of state-of-the-art deep learning models relies heavily on vast datasets, often sourced from \textit{free-to-use} online platforms without explicit consent. This unauthorized acquisition poses significant risks, as seen in cases where personal data, including facial images and medical records, have been exploited in commercial AI models~\cite{hill2020secretive,arstechnicaArtistFinds}. Public concern about these practices has spurred interest in privacy protections, reinforced by regulations such as GDPR~\cite{voigt2017eu} and CCPA~\cite{illman2019california}. However, these regulations focus primarily on data collection and governance rather than addressing the core issue: once data are acquired and integrated into AI training pipelines, few mechanisms exist to prevent its continued use. To bridge this gap, recent research has introduced the concept of making data unlearnable to AI models. Huang \textit{et al.}~\cite{huang2021unlearnable} proposed unlearnable examples (UE), a technique that adds imperceptible noise to the data, causing models to fail in extracting useful information. This approach uses \textit{error-minimizing} (EM) noise to yield near-zero loss during training, tricking the model into ``learning'' nothing from the protected data.

Built on this foundation, subsequent research has focused on enhancing the robustness of UEs. For instance, Fu \textit{et al.}~\cite{fu2022robust} proposed robust unlearnable examples (REM), resilient to adversarial training techniques that could otherwise make standard UEs learnable. Similarly, Liu \textit{et al.}~\cite{liu2024stable} further improved the stability of these robust UEs by designing defensive noise that resists random perturbations, thus eliminating the need for complex adversarial perturbation. Expanding on this idea, Ren \textit{et al.}~\cite{ren2022transferable} enhanced the transferability of error-minimizing noise using the Classwise Separability Discriminant to maximize model confusion across different data categories. Moreover, application-specific adaptations of unlearnable examples have emerged, targeting areas such as diffusion models~\cite{zhao2023unlearnable}, time series data~\cite{jiang2024unlearnable}, and medical image segmentation~\cite{lin2024safeguarding}. Yet, despite their promise, UEs still struggle with two critical challenges that hinder their effectiveness:

\noindent\textbf{(1) Fragility of UEs:} Recent studies have exposed significant vulnerabilities in the robustness of UEs. Existing methods rely on error-minimizing noise to render images uninformative to AI models. However, researchers have proposed different relearning strategies based on adversarial training \cite{tao2021better} or image transformations like resizing, compression, and grayscale conversion \cite{liu2021going,qin2023learning}. These methods can significantly weaken the effectiveness of error-minimization noise, rendering these defenses unreliable for long-term data security. 

\noindent\textbf{(2) Poor Usability for Real-World Applications:}
While some approaches have attempted to enhance the robustness of UEs by introducing adversarial perturbation~\cite{fu2022robust} or by stabilizing the UEs using random perturbation~\cite{liu2024stable}, they often lead to significant image degradation as unlearnability increases. Stronger noise improves protection but makes images unusable for photographers, artists, and social media users who rely on high-quality visuals. Grainy, artifact-ridden images are impractical for professional work and unappealing for casual sharing. It is evident from Fig.~\ref{fig: utility_vs_usability} that the robustness comes with the cost of perceptual quality for these methods. Moreover, pixel-space noise disrupts format consistency, leading to compression artifacts and post-processing errors.

\textbf{\methodName.} 
To address the aforementioned limitations, we propose \methodName, a robust defense framework that achieves greatly improved trade-offs between data perceptual quality and unlearnability. Our method is particularly effective against various state-of-the-art data relearning strategies, such as adversarial training/augmentations~\cite{fu2022robust}, grayscale conversion~\cite{liu2021going}, data purification~\cite{jiang2023unlearnable}, and image transformation~\cite{qin2023learning}.
Unlike conventional methods that apply pixel-space noise, \methodName operates in the semantic space, leveraging a pre-trained diffusion autoencoder~\cite{preechakul2022diffusion}. Specifically, it encodes an input image $x$ into a high-level semantic representation $z_{\text{sem}}$ and a stochastic code $x_T$, then optimizes a defensive semantic noise $z_{\delta}$ to make the image unlearnable. The final image $x_u$ is reconstructed through a Conditional Denoising Diffusion Implicit Model (DDIM)~\cite{song2020denoising} decoding process. 
This optimization process ensures that $x_u$ induces a near-zero loss in a surrogate model, ensuring that models fail to extract meaningful features.
To maintain both robustness and visual quality, we conduct empirical studies to determine the noise radius $\rho_u$~
ensuring that semantic space perturbations translate into effective pixel domain disruptions, changing the image's intrinsic features. By aligning the image's intrinsic features with model gradient space, \methodName enhances resistance against adversarial retraining, image transformations, and lossy compression. Unlike unnatural, high-frequency ``static'' artifacts~\cite{zhu2023frequency} introduced by pixel-based noise, our method guides semantic modifications along the learned image manifold, creating coherent variations such as gentle lighting shifts or subtle tone adjustments. These semantic changes align closely with human perception, appearing natural~\cite{wang2023generating}.

Extensive experiments with multiple UE baselines, various relearning strategies, and diverse settings across four datasets demonstrate the superior effectiveness of our system, particularly in challenging relearning scenarios.

\vspace{-3mm}
\section{Related Work}
\vspace{-2mm}

\noindent\textbf{Unlearnable Examples.} 
Unlike poisoning attacks~\cite{biggio2012poisoning,koh2017understanding,shafahi2018poison}, which degrade model performance by injecting malicious data, unlearnable examples (UEs)~\cite{huang2021unlearnable} defensively add imperceptible perturbations to render data uninformative for training, thereby limiting generalization. Early work explored diverse UE mechanisms, including channel/appearance manipulations~\cite{liu2021going} and robustness to adversarial training via Robust Unlearnable Examples (REM)~\cite{fu2022robust}. Subsequent efforts improved transferability and stability across models and training pipelines~\cite{ren2022transferable,liu2024stable} and extended UEs to generative diffusion models~\cite{zhao2023unlearnable}.

More recent work has shifted toward stronger robustness, guarantees, and broader modalities. Meng \textit{et al.}~\cite{meng2024semantic} proposed semantic Deep Hiding using invertible neural networks to produce more resilient UE perturbations, while Zhu \textit{et al.}~\cite{zhu2024detection} studied the detectability of UEs and corresponding defenses, highlighting a growing arms race. Wang \textit{et al.}~\cite{wang2025provably} provided a certification framework and constructed provably unlearnable examples, offering formal learnability guarantees. Beyond images, Meerza \textit{et al.}~\cite{meerza2025harmonycloak} introduced HarmonyCloak, extending unlearnable examples to the music domain by leveraging psychoacoustic masking and semantic guidance to protect audio content from generative music models while preserving perceptual quality. Wu \textit{et al.}~\cite{wu2025temporal} introduced Temporal Unlearnable Examples for protecting personal video data against tracking-based exploitation, and Li \textit{et al.}~\cite{li2025versatile} investigated versatile transferable UE generators across challenging shifts (e.g., architectures and resolutions). Despite these advances, achieving strong robustness against adaptive retraining while preserving perceptual quality and maintaining broad real-world effectiveness remains challenging.

\noindent\textbf{Relearning from UEs.}
Recent research has explored countermeasures that attempt to restore learnability to UEs~\cite{dolatabadi2024devil,jiang2023unlearnable,qin2023learning}. One of the earliest approaches by Tao \textit{et al.}~\cite{tao2021better} showed that Adversarial Training (AT) can weaken UE protections by introducing adversarial perturbations during training. Building on this idea, Qin \textit{et al.} proposed UEraser~\cite{qin2023learning}, which combines data augmentation with adversarial training to improve model generalization on protected data. Other methods adopt alternative strategies. Liu \textit{et al.}~\cite{liu2021going} demonstrated that grayscale transformations can neutralize channel-specific UE noise, improving model robustness. Diffusion-based purification has also gained attention: Jian \textit{et al.} proposed LUE~\cite{jiang2023unlearnable}, which uses joint-conditional diffusion to restore learnability, while Dolatabadi \textit{et al.} introduced \textsc{Avatar}~\cite{dolatabadi2024devil}, a diffusion-based approach that removes defensive noise from UEs. These countermeasures highlight the ongoing arms race between privacy-preserving defenses and adaptive learning strategies, underscoring the need for more resilient UE techniques.

\vspace{-3mm}
\section{Preliminaries}
\vspace{-2mm}
\subsection{Diffusion Autoencoders}
\vspace{-1mm}

Diffusion autoencoders unify the strengths of denoising diffusion probabilistic models (DPMs)~\cite{ho2020denoising} and Denoising Diffusion Implicit Models (DDIMs)~\cite{song2020denoising} to enable high-fidelity image reconstructions from semantically structured latent codes. DPMs progressively corrupt images with Gaussian noise and use a U-Net denoiser to learn the reverse process, while DDIMs offer a faster, deterministic variant.

Building on this foundation, diffusion autoencoders~\cite{preechakul2022diffusion} introduce a semantically structured latent space that enhances controllability and interpretability. The encoder decomposes an image $x_0$ into two latent variables: a high-level semantic code $z_{sem} = \mathcal{E}(x_0)$ and a low-level stochastic latent $x_T$. The semantic code $z_{sem}$, a compact 512-dimensional vector, captures global, non-spatial semantics akin to the style vector in StyleGAN~\cite{karras2019style}, while $x_T$ preserves local and textural details via reverse DDIM encoding. During decoding, the model reconstructs the image using a DDIM-based generative process conditioned on both $z_{sem}$ and $x_T$:
{\small
\begin{equation}
\setlength{\abovedisplayskip}{3pt}
\setlength{\belowdisplayskip}{3pt}
p_\theta(x_{0:T} | z_{sem}) = p(x_T) \prod_{t=1}^T p_\theta(x_{t-1} | x_t, z_{sem}),
\end{equation}
}
where $p_\theta(x_{t-1} | x_t, z_{sem}) = \mathcal{N}(f_\theta(x_t, t, z_{sem}), 0)$ for $t=1$, and follows the DDIM update for other $t$. The conditional mean function $f_\theta$ integrates the semantics $z_{sem}$ to guide denoising, enabling the model to generate coherent and diverse outputs consistent with the encoded structure.

This dual-latent formulation, comprising semantic $z_{sem}$ and stochastic $x_T$, enables disentangled control over content and style, supporting tasks such as semantic interpolation, content-preserving manipulation, and reconstruction, while preserving the high-fidelity synthesis capabilities enabled by diffusion models.

\vspace{-2mm}
\subsection{Unlearnable Examples}
\vspace{-1mm}
Conventionally, the process of crafting UEs for classification models involves modifying each data sample to make the model’s \textit{classification loss} as close to zero as possible. This \textit{zero-loss} approach essentially tricks the model into perceiving these samples as trivial or uninformative, thereby suppressing any gradients that could guide effective learning~\cite{huang2021unlearnable}. 
To achieve this, consider a data sample $x$ with label $y$, the defender can craft \textit{error-minimizing} noise $\delta$ that, when added to $x$, minimizes the model's learning potential for this sample. The noise $\delta$ is generated by solving a bi-level optimization problem:
{\small
\begin{equation} 
\setlength{\abovedisplayskip}{3pt}
\setlength{\belowdisplayskip}{3pt}
  \arg \underset{\theta}{\min} \quad \mathbb{E}_{x,y}[\underset{\delta}{\min}\mathcal{L}(g_\theta(x+\delta),y)] \quad s.t.\quad \|\delta\|_p \leq\rho,
\label{eq2}
\end{equation}
}
where $g_{\theta}$ denotes the classification model, $\mathcal{L}$ is the cross-entropy loss, and the noise magnitude $\|\delta\|$ is bounded by $\rho$.

\begin{figure}[t]
    \centering
    \includegraphics[width=0.9\linewidth]{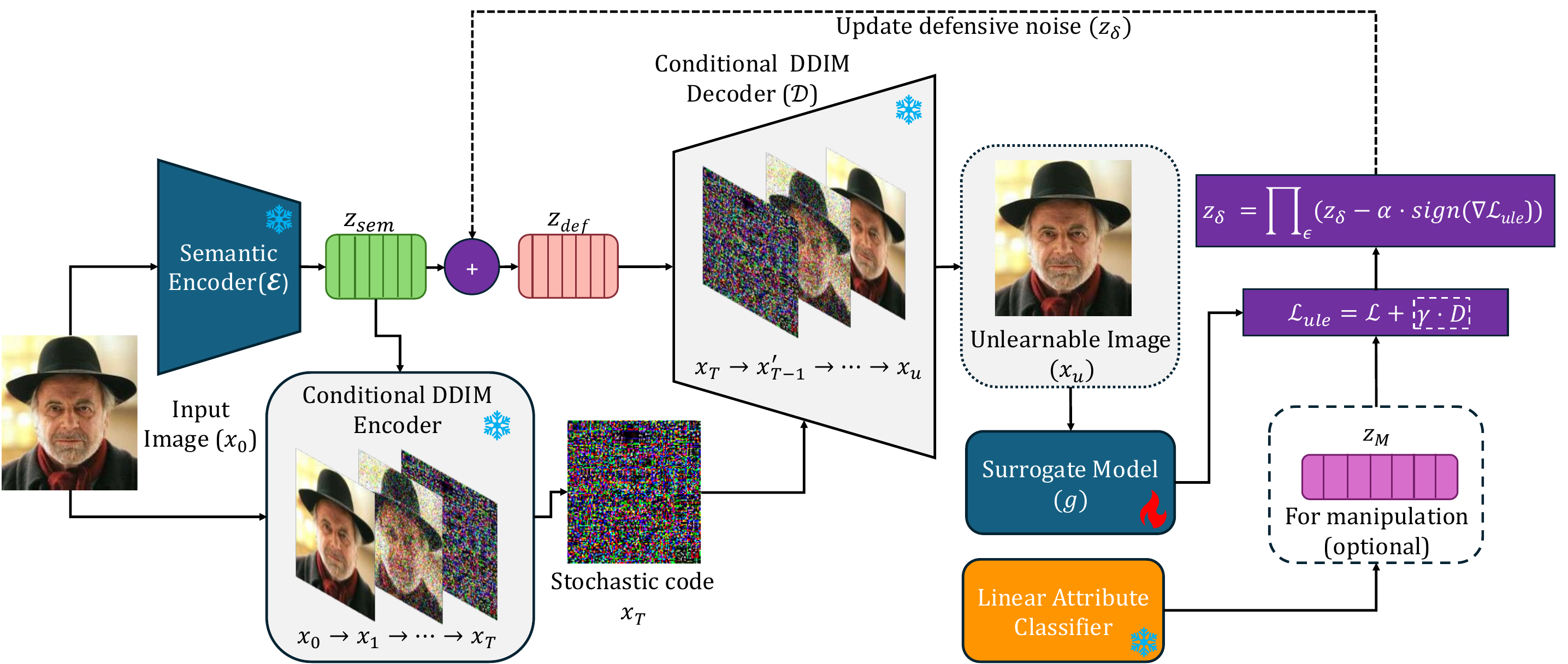}
    \vspace{-1mm}
    \caption{Overview of \methodName: \methodName leverages a diffusion autoencoder to encode the input image into a semantic code $z_{sem}$ and a stochastic code $x_T$. It then iteratively optimizes a defensive semantic code $z_{\delta}$ so that the resulting unlearnable image $x_u$, generated by the conditional DDIM decoding process, tricks the surrogate model $g_\theta$ into believing there is nothing to learn.
    }
    \vspace{-7mm}
    \label{fig: framework}
\end{figure}

\vspace{-3mm}
\section{Design of\textbf{~\methodName}}
\vspace{-2mm}
\subsection{Overview of \textbf{\methodName}}
\vspace{-1mm}
The overall pipeline of \methodName is depicted in Fig. \ref{fig: framework}. \methodName utilizes a diffusion-based autoencoder architecture~\cite{preechakul2022diffusion}, comprising a semantic encoder $\mathcal{E}$ and a conditional DDIM, which contains a stochastic encoder and an image decoder $\mathcal{D}$. In this process,  an input image $x_0$ is encoded into a high-level semantic representation $z_{sem}$ and a stochastic code $x_T$ that captures low-level variations. The objective is to optimize a defensive semantic noise $z_{\delta}$ such that the perturbed semantic representation $z_{def} = z_{sem} + z_{\delta}$ produces an unlearnable image $x_u$, causing the surrogate model $g$ to experience near-zero loss. Given $x_0$, we can generate the defensive noise $z_{\delta}$ by solving the bi-level optimization problem:
{\small
\begin{equation}
\setlength{\abovedisplayskip}{3pt}
\setlength{\belowdisplayskip}{3pt}
\begin{split}
    &\arg \underset{\theta}{\min} \; \mathbb{E}_{(x_0,y)} \left[\underset{z_{\delta}}{\min}\;\mathcal{L} (g_\theta (\mathcal{D}(\mathcal{E}(x_0)+ z_{\delta}, x_T)), y)\right],\\
    &\qquad\qquad\qquad \qquad \qquad \qquad\qquad s.t. ||z_{\delta}||_p \leq \rho_u,
\label{equ:main_loss}
\end{split}
\end{equation}
}
where $z_{sem} = \mathcal{E}(x_0)$, $x_u = \mathcal{D}(\mathcal{E}(x_0)+ z_{\delta}, x_T)$ and $\rho_u$ is the noise bound. This formulation is a min-min bi-level optimization problem where the inner minimization seeks to find $z_{\delta}$, constrained by an $L_p$-norm, that minimizes the model's classification loss, while the outer minimization optimizes the model parameters $\theta$ to further reduce the loss. 

To generate effective unlearnable examples, it is crucial to understand the model's learning dynamics, specifically the gradient trajectory of parameters $\theta$ during training. Thus, $\theta$ must be partially trained initially to reflect the model’s typical behavior. This enables $\delta$ to be optimized in alignment with learning dynamics, guiding the model toward low training loss. Unlike adversarial examples, which induce prediction failures, here $\delta$ is crafted to create the illusion of successful learning. Specifically, we optimize $\theta$ for $Q$ steps before updating $\delta$. This process repeats until the model achieves an error rate below a predefined threshold $\lambda$, terminating the training.

Additionally, the conditional DDIM is conditioned on the defensive semantic code $z_{def} = z_{sem} + z_{\delta}$, with the noise prediction network $f(x_T, T, z_{def})$ estimating noise at each step, where $T$ indicates the number of noise addition and removal steps. During the decoding phase, the deterministic generative process reconstructs the final image $x_u$ based on the conditioned input.

\vspace{-4mm}
\subsection{Defensive Noise in the Semantic Space}
\vspace{-1mm}
In \methodName, we focus on generating defensive noise within the semantic space of diffusion autoencoders. To achieve this, we inject noise into the semantic space using the first-order optimization method, Projected Gradient Descent (PGD)~\cite{madry2018towards}. PGD works by iteratively adjusting the input in the direction of the gradient of a loss function, ensuring that each step stays within a defined perturbation limit. The constrained inner minimization problem is solved with the following update rule:
{\small
\begin{equation}
\setlength{\abovedisplayskip}{3pt}
\setlength{\belowdisplayskip}{3pt}
    z_{\delta(t+1)} = \Pi_{\rho_u} (z_{\delta(t)} - \eta \cdot \mathrm{sign}(\nabla_{z_\delta} \mathcal{L}_{ule})),
\end{equation}}
where $\mathcal{L}_{ule}$ represents the loss function defined as $\mathcal{L}_{ule} = \mathcal{L}(g_{\theta}(\mathcal{D}(\mathcal{E}(x_0) + z_{\delta}, x_T)), y)$, where $t$ is the current step of the perturbation, and $\eta$ is the step size. The function $\Pi_{\rho_u}$ is a projection function that ensures the noise $z_{\delta}$ remains within the allowed perturbation range $\rho_u$ by clipping it if it exceeds the limit. In our framework, PGD iteratively refines the noise $z_{\delta}$, to hinder the surrogate model $g_{\theta}$'s learning.

\noindent\textbf{Benefits of Injecting Semantic Noises.}
Semantic perturbations offer a unique advantage over pixel-level noise by creating subtle adjustments that alter an image’s underlying attributes without introducing detectable spatial patterns. Human perception, which processes high-level attributes non-linearly, can naturally integrate and overlook these subtle semantic changes, making them imperceptible~\cite{wang2023generating}. In contrast, pixel-space perturbations introduce noticeable patterns perceived as foreign elements~\cite{song2018constructing}. Neural networks, however, respond sensitively even to minor semantic adjustments, directly impacting the linear feature space of model predictions. Thus, effective unlearnable behavior can be achieved with a smaller perturbation budget, reducing pixel-space distortion and allowing our method to outperform previous approaches in visual quality and robustness.

Moreover, semantic perturbations inherently improve robustness by altering high-level attributes instead of pixel-level noise. This makes them less vulnerable to pixel-domain augmentations~\cite{song2018constructing} and filtering techniques~\cite{xie2019feature} commonly used to counteract defensive effects. Since these semantic changes do not present fixed spatial patterns, isolating or removing them becomes considerably more challenging. In the domain of adversarial examples~\cite{xiao2018generating}, photorealistic perturbations that closely align with the natural characteristics of image objects have proven to be resilient to adversarial training. Using semantic-space transformations, \methodName improves the trade-off between perceptual quality and unlearnability while enhancing robustness against pixel-space attacks.

\vspace{-2mm}
\subsection{UEs via Controlled Modifications} 
\vspace{-1mm}
Adding noise in the semantic space allows for subtle, meaningful adjustments to an image that is generally more acceptable to human observers than pixel-level noise. This enables controlled modifications, such as attribute or style adjustments, allowing users to share visually enhanced images online. In this work, we focus on two modification types: attribute manipulation (e.g., adding a smile) and style manipulation (e.g., adjusting tone or contrast), where defensive noise $z_{\delta}$ alters specific image features.

By manipulating semantic features, our approach can also leverage targeted semantic directions to create purposeful, visible modifications, ensuring both unlearnability and perceptual changes.
To achieve this, we adjust the defensive semantic code $z_{def}$ in the direction of the desired modification, determined by the weight vector of a linear classifier trained to distinguish positive and negative instances of the attribute or style. For example, to add a smile, we use a classifier that takes semantic features of an image as input and distinguishes between smiling and non-smiling faces. The classifier’s weight vector, which captures the relationship between the semantic features associated with smiling, is then used to guide the defensive noise in that direction. Similarly, for style changes, such as applying certain filter effects, the classifier weights capture the characteristic patterns of these effects in semantic space. The modified loss function incorporating these controlled changes for effective unlearnability is as follows:
{\small
\begin{equation}
\setlength{\abovedisplayskip}{3pt}
\setlength{\belowdisplayskip}{3pt}
    \mathcal{L}_{ule} = \mathcal{L}(g_\theta(\mathcal{D}(\mathcal{E}(x_0) + z_{\delta}, x_T)), y) + \gamma \cdot (1-cos(z_{\delta}, z_{M})),
\end{equation}}
where $z_{M}$ is the desired semantic direction calculated by normalizing the weight vector of the relevant classifier $w_{cls}/||w_{cls}||$, $cos(.)$ denotes the cosine similarity, $\gamma$ is a scaling factor that modulates the degree of desired modifications, and the cosine similarity is defined as $D$ in Fig.~\ref{fig: framework}. Cosine similarity aligns the noise vector in the target direction (direction of the semantics of the desired effect $z_{M}$), allowing for controlled, intentional adjustments within the semantic space. This approach enhances practical usability while maintaining resistance to unauthorized usage.

\vspace{-3mm}
\section{Experiments}
\vspace{-2mm}
\subsection{Experimental Setup}
\vspace{-1mm}
\label{sec.setup}
\textbf{Datasets and Model Architecture.} We conducted experiments on three major vision benchmark datasets: CIFAR-10, CIFAR-100 \cite{krizhevsky2009learning}, and the first 100 classes of ImageNet \cite{russakovsky2015imagenet}. For data augmentation, we applied random cropping and horizontal flipping. For the experiment of controlled modifications, we used the CelebA-HQ dataset \cite{karras2017progressive}, and ImageNet \cite{russakovsky2015imagenet}. The proposed method and baselines were evaluated across various vision tasks using ResNet18 \cite{he2016deep} as a surrogate model. We use a pre-trained DiffAE~\cite{preechakul2022diffusion} model trained on different datasets. Following prior UE works, DiffUE employs an iterative bi-level optimization procedure with 10 PGD inner steps per surrogate model update, and optimization continues until the surrogate model reaches over 99\% training accuracy.

\noindent\textbf{UE Baselines.} We compare \textbf{\methodName} with state-of-the-art approaches for generating UEs. Specifically, we evaluate against error-minimizing noise {(\textbf{EM})}~\cite{huang2021unlearnable}, which is the first work in the field of UEs and directly applies error minimization noise to the images. We also include robust error-minimizing noise {(\textbf{REM})}~\cite{fu2022robust}, which adds noise designed to reduce adversarial training loss, and stable error-minimizing noise {(\textbf{SEM})}~\cite{liu2024stable}, which generates unlearnable noise that is resilient to random perturbations. These baselines represent the foundational UE generation paradigms widely adopted in prior literature, and many subsequent methods build upon or extend these error-minimization frameworks.

\noindent\textbf{Techniques Used for Robustness Evaluation.} To assess the robustness of our method, we apply Mean and Gaussian filters to test the resilience of our images against noise-filtering approaches. We evaluate our approach using recent state-of-the-art relearning techniques designed to counter UEs and restore their learnability. Specifically, we tested \textbf{Adversarial Training}~\cite{tao2021better}, which introduces adversarial noise to bypass defensive noise and recover learnability in UEs. We also used \textbf{GrayScale}~\cite{liu2021going}, a method that converts UEs to grayscale to neutralize color-channel noise, \textbf{UEraser}~\cite{qin2023learning}, which employs adversarial augmentations to reintroduce learnability, as well as \textbf{LUE}~\cite{jiang2023unlearnable}, which uses a diffusion purification model to eliminate unlearnable noise from images. Additionally, we evaluate the effectiveness of our approach under common lossy image compression formats, such as JPEG, WebP, and HEIC. 

\begin{figure}[t]
    \centering
    \includegraphics[width=0.7\linewidth]{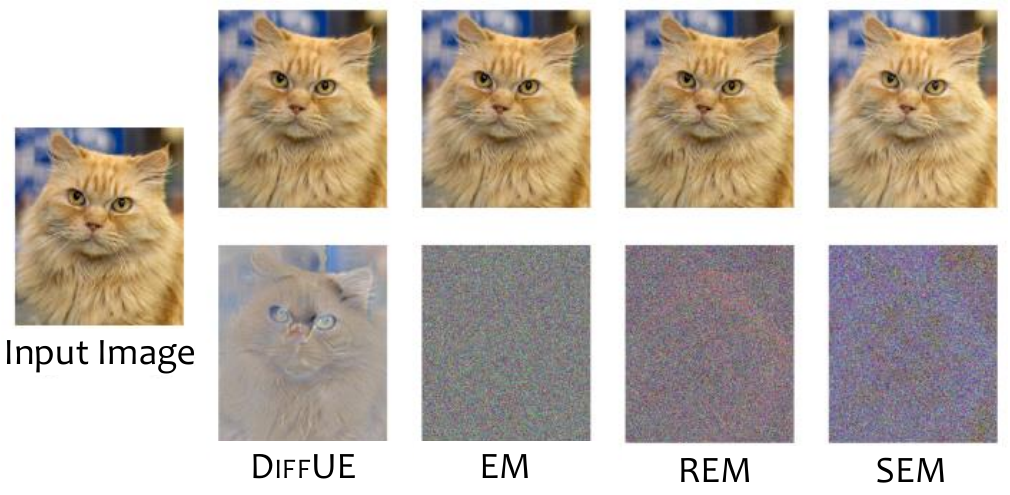}
    \vspace{-4mm}
    \caption{Visualization of noise types and crafted examples for the ImageNet Subset.}
    \vspace{-6mm}
    \label{fig: noise_comparison}
\end{figure}

\vspace{-2mm}
\subsection{Semantic Noise Radii Selection}
\vspace{-1mm}
Unlike existing UE methods that constrain noise levels in the pixel space (e.g., 2/255, and 4/255), \methodName introduces defensive noise in the semantic space, where the noise boundary (i.e., $\rho_u$ in Equation~\ref{equ:main_loss}) cannot be directly mapped to the pixel space due to non-linear transformations between the two spaces. Moreover, defensive noise in the semantic space generates more photorealistic pixel-level changes, as illustrated in Fig.~\ref{fig: noise_comparison}. Even with the same amount of pixel modifications, \methodName can more effectively conceal perturbations within the original images. Therefore, instead of relying on pixel-level perturbation bounds to assess perceptual impact, as in existing studies, we use widely adopted metrics to quantify image perceptual quality (see section~\ref{subsec:quality} for details).

To ensure the semantic noise radius is set to a reasonable value, we empirically measure the maximum pixel changes induced by a given noise level for each dataset. Specifically, we perturb 50 of the 512 dimensions of the semantic code and analyze their impact on the reconstructed image in the pixel space. Based on the analysis on 2,000 samples per dataset, we set the semantic noise radius to 0.20 for CIFAR-10/100 and 0.35 for ImageNet, corresponding to a maximum pixel shift of 4/255.

While increasing the noise radius can further enhance unlearnability, it may also distort the image’s semantics. Although the utility-unlearnability trade-off could be further optimized through fine-tuning, we did not adjust the radius further, as our chosen setting already achieves significantly stronger unlearnability and robustness against relearning, even when compared to baselines with perturbation radius of 8/255, while also achieving higher perceptual quality.

 \begin{table}[t]
\centering
\caption{Test accuracy (\%) of models trained on data protected by different noise strategies under adversarial training with varying perturbation radii. The defensive perturbation radius $\rho_u$ is fixed for all the methods. The training adversarial perturbation radius $\rho_a$ and $\rho_r$ for REM and SEM are adjusted. A lower test accuracy indicates better protective effectiveness.} \vspace{2mm}

\resizebox{0.87\linewidth}{!}{
\centering
\begin{tabular}{ccccccccccc}
\toprule
 \multirow{2}{*}{Dataset} &
  \multirow{2}{*}{Adv.Train.} &
  \multirow{2}{*}{Clean} &
  \multirow{2}{*}{EM} &
  \multirow{2}{*}{\methodName} &
  \multicolumn{3}{c}{REM} &
  \multicolumn{3}{c}{SEM} \\ \cmidrule(l){6-11} 
   &$\rho_t$
   &
   &
   &
   &
  $\rho_a=$ 0 &
  2/255 &
  4/255 &
  $\rho_r=$ 0 &
  2/255 &
  4/255 \\ \midrule
\multirow{3}{*}{CIFAR-10} &0 &95.64 &11.98 &\textbf{10.79} &13.22 &23.91 &31.27 &18.92 &18.51 &25.14\\
 &2/255 &89.24 &20.96 &\textbf{12.59} &68.54 &35.37 &37.75 &20.82 &23.97 &26.65 \\
 &4/255 &86.92 &76.95 &\textbf{17.11} &86.55 &75.47 &64.64 &73.57 &46.19 &46.85 \\ 
 \midrule

\multirow{3}{*}{CIFAR-100} & 0 &85.32 &\textbf{1.76} &2.17 &2.11 &9.82 &19.3 &9.81 &4.09 &18.61 \\
 &2/255 &76.76 &67.29 &\textbf{3.78} &51.81 &10.44 &18.35 &52.52 &10.44 &12.56 \\
 &4/255 &72.19 &64.59 &\textbf{4.92} &65.93 &61.29 &32.22 &60.64 &47.52 &26.19 \\ 
 \midrule
\multirow{3}{*}{ImageNet Subset} &0 &82.54 &\textbf{1.89} & 2.11 &3.45 &9.13 &17.74 &4.15 &8.14 &10.11\\
 &2/255 &73.22 &70.67 &\textbf{12.45} &48.56 &30.22 &31.40 &35.21 &33.22 &36.22\\
 &4/255 &70.23 &67.86 &\textbf{17.20} &65.58 &57.23 &45.49 &57.32 &38.11 &38.57\\ \bottomrule
\end{tabular}}
\vspace{-4mm}
\label{table: unlearnability}
\end{table}

\begin{table}[t]
    \centering
    \caption{ Test accuracy (\%) of various methods trained on the unlearnable CIFAR-10 dataset, evaluated under different filtering techniques and relearning strategies.}\vspace{2mm}
    \label{tab: attacks}
    \resizebox{0.55\linewidth}{!}{
        \begin{tabular}{@{}ccccccc@{}}
            \toprule
            \multirow{2}{*}{\textbf{Method} } &\multicolumn{2}{c}{Filters} &\multicolumn{3}{c}{ Relearning strategies}\\
            &Mean &Gaussian &GrayScale & UEraser & LUE \\ \midrule
            \textbf{Clean}   &90.65 &91.26 &91.65 &92.78 &91.17 \\
            \textbf{EM}      &37.87 &32.71 &83.23 &90.23 &90.33 \\
            \textbf{REM}     &29.92 &28.53 &63.87 &87.39 &89.57\\
            \textbf{SEM}     &28.55 &24.22 &34.23 &81.56 &89.29\\
            \textbf{\methodName}  &\textbf{13.32} &\textbf{12.43} &\textbf{13.99} &\textbf{14.19} &\textbf{33.97}\\ \bottomrule
        \end{tabular}
    }
    \vspace{-5mm}
\end{table}

\begin{table}[t]
    \centering
     \caption{Impact of image compression on different defensive noise evaluated on different datasets. Table shows the test accuracy (\%) of various methods.}\vspace{2mm}
    \resizebox{0.75\linewidth}{!}{
        \begin{tabular}{@{}c|cccc|cccc@{}}
            \toprule
            \multirow{2}{*}{\textbf{Method}}  &\multicolumn{4}{c|}{CIFAR-10}&\multicolumn{4}{c}{ImageNet}\\
                             &No Comp. &JPEG & WebP & HEIC &No Comp.&JPEG & WebP & HEIC\\ \midrule
            \textbf{EM}      &11.98 &45.22 &55.57 &35.62 &\textbf{1.89} &34.21 &41.98 &37.57\\
            \textbf{REM}     &13.22 &38.12 &41.78 &34.95 &3.45 &33.45 &43.22 &37.12\\
            \textbf{SEM}     &18.92 &32.92 &35.96 &23.89 &4.15 &18.33 &29.45 &17.22\\
            \textbf{\methodName}  &\textbf{10.79}&\textbf{17.22} &\textbf{18.05} &\textbf{18.11} &2.11 &\textbf{4.12} &\textbf{7.88} &\textbf{5.11}\\ \bottomrule
        \end{tabular}
    }
    \label{tab: image_compression}
    \vspace{-3mm}
\end{table}

\vspace{-2mm}
\subsection{Effectiveness of~\textbf{\methodName}}
\vspace{-1mm}
We applied various UE methods to the entire training set, setting the defensive perturbation radius $\rho_u$ to $8/255$ for all baselines. For \methodName, semantic noise radii ($\rho_u$) were determined empirically for each dataset: $0.2$ for CIFAR-10 and CIFAR-100, and $0.35$ for the ImageNet Subset, targeting a maximum pixel-level change of $4/255$, corresponding to a $50\%$ smaller noise budget than the baselines. REM and SEM jointly optimize defensive perturbations with an additional noise component, namely adversarial perturbation (radius $\rho_a$) in REM and random perturbation (radius $\rho_r$) in SEM. We evaluate different $\rho_a$ and $\rho_r$ values during unlearnable image generation and subsequently perform adversarial training with varying adversarial radii $\rho_t$. Table~\ref{table: unlearnability} reports the resulting test accuracies on clean data. As $\rho_t$ increases, models trained on baseline unlearnable datasets achieve higher test accuracy, indicating that stronger adversarial training makes these methods less effective. For example, with $\rho_t = 4/255$, test accuracy reaches $76.95\%$, $64.64\%$, and $46.85\%$ for EM, REM ($\rho_a=4/255$), and SEM ($\rho_r=4/255$), respectively, on CIFAR-10. In contrast, \methodName maintains a low accuracy of $17.11\%$, demonstrating greater resistance to adversarial training. A similar trend appears on the ImageNet Subset, where \methodName remains at $17.20\%$ while REM and SEM increase to $45.49\%$ and $38.57\%$, respectively. Fig.~\ref{fig: noise_comparison} illustrates the generated perturbations. Unlike pixel-based noise, \methodName produces perceptually realistic perturbations that preserve visual fidelity while making the protective signal more difficult for adversarial training and filtering methods to remove.

\vspace{-2mm}
\subsection{Robustness Evaluation}
\vspace{-1mm}
We further evaluate the robustness of \methodName and baseline methods against real-world transformations that could compromise data protection.
Table~\ref{tab: attacks} compares the resilience of various unlearnable noise methods under different filtering techniques and relearning strategies. The table shows that \methodName achieves the lowest test accuracy across all filtering techniques (Mean: 13.32\%, Gaussian: 12.43\%) and remains more resistant to relearning strategies such as UEraser (14.19\%) and LUE (33.97\%), compared to the next-best method (SEM: 81.56\% under UEraser, 89.29\% under LUE). This highlights the robustness of \methodName against pixel-domain filtering and augmentation.

Given that most images undergo lossy compression before online sharing, we also assess the impact of compression on unlearnable noise in Table~\ref{tab: image_compression}. \methodName exhibits minimal degradation when compressed with JPEG (17.22\% CIFAR-10, 4.12\% ImageNet) or WebP (18.05\% CIFAR-10, 7.88\% ImageNet), whereas pixel-based baselines degrade significantly (e.g., EM: 45.22\% CIFAR-10, 34.21\% ImageNet for JPEG). By avoiding pixel-space noise, \methodName maintains unlearnability under real-world transformations.

\begin{table}[t]
\centering
\caption{Comparison of perceptual metrics with state-of-the-art methods on CelebA-HQ and ImageNet. The defensive perturbation radius for \methodName is set to $\rho_u=0.25$ for CelebA-HQ and $\rho_u=0.35$ for ImageNet, while EM, REM, and SEM use a perturbation radius of $\rho_u=8/255$ and $\rho_a = 4/255$ for REM and $\rho_r = 4/255$ for SEM.}
\vspace{2mm}
\resizebox{0.6\linewidth}{!}{

    \begin{tabular}{@{}lccccc@{}}
    \toprule
    \textbf{Dataset} &\textbf{Method} & EM     & REM    & SEM    & \methodName \\ \midrule
    \multirow{4}{*}{CelebA-HQ}       &\textbf{FID}($\downarrow$)   & 11.612 & 10.971 & 10.145 & \textbf{5.355} \\ 
                                     &\textbf{SSIM}($\uparrow$)    & 0.704  & 0.561  & 0.611  & \textbf{0.814} \\
                                     &\textbf{PSNR}($\uparrow$)    & 30.112 & 29.411 & 29.871 & \textbf{32.438}\\
                                     &\textbf{Test Acc.} ($\downarrow$) &55.98\% &54.27\% &52.14\% & \textbf{50.22\%}\\
    \midrule     
    \multirow{3}{*}{ImageNet}        &\textbf{FID}($\downarrow$)   &10.583  &11.226  &11.874 &\textbf{5.492}  \\ 
                                     &\textbf{SSIM}($\uparrow$)    & 0.682  & 0.389  & 0.576  & \textbf{0.781} \\
                                     &\textbf{PSNR}($\uparrow$)    & 29.091 & 23.295 & 29.044 & \textbf{32.934}\\
                                     &\textbf{Test Acc.} ($\downarrow$) &\textbf{1.89}\% &17.74\%&10.11\% & 2.11\%\\
    \bottomrule
    \end{tabular}
}
\label{tab: perceptual-qualities}
\vspace{-5mm}
\end{table}

\vspace{-2mm}
\subsection{Perceptual Quality Evaluation}
\vspace{-1mm}
\label{subsec:quality}
We argue that UEs should maintain high perceptual quality so users can use them without compromises. To assess perceptual quality, we use three metrics: Frechet Inception Distance (FID)~\cite{heusel2017gans} for naturalness, Structural Similarity Index Measure (SSIM)~\cite{wang2017structural} for similarity to the clean image, and Peak Signal-to-Noise Ratio (PSNR)~\cite{instruments2013peak} for image quality under noise. Table~\ref{tab: perceptual-qualities} compares unlearnable examples generated by various methods on CelebA-HQ and ImageNet. On CelebA-HQ, \methodName achieves the lowest FID (40–50\% reduction) with the highest PSNR (32.438) and the highest SSIM (0.814), indicating natural unlearnable examples with low perceptual noise. On ImageNet, \methodName shows similar trends, with a 40–50\% FID improvement and higher SSIM and PSNR than baselines.

\begin{figure}[t]
    \centering
    \begin{subfigure}[t]{0.38\linewidth}
        \centering
        \includegraphics[width=\textwidth]{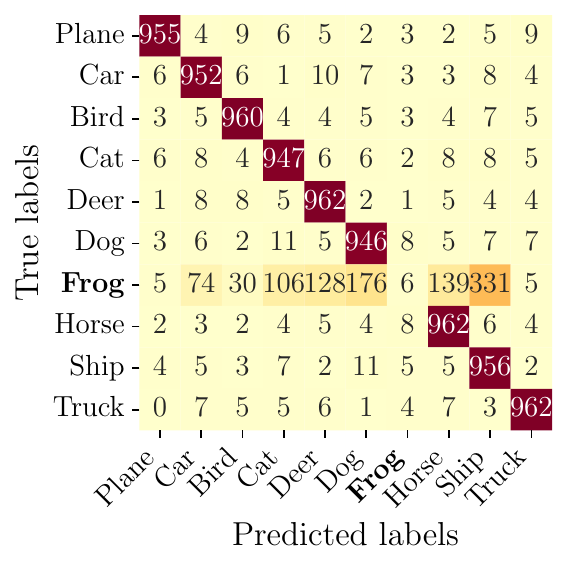}
        \caption{Single Class}
        \label{single_class}
    \end{subfigure}%
    \hfill
    \begin{subfigure}[t]{0.38\linewidth}
        \centering
        \includegraphics[width=\textwidth]{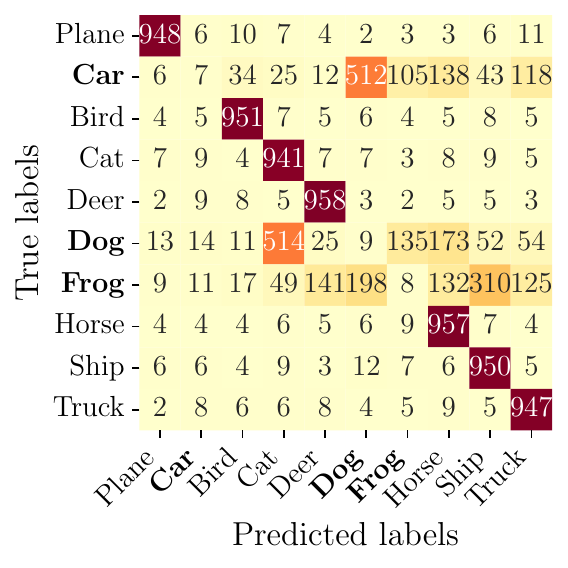}
        \caption{Multiple Classes}
        \label{multi_class}
    \end{subfigure}
    \vspace{-2mm}
    \caption{Classwise unlearnability analysis. In (a), we have perturbed only one class and kept the other classes clean, and in (b), we have perturbed multiple classes.}
    \vspace{-5mm}
    \label{fig: Classwise}
\end{figure}

\vspace{-2mm}
\subsection{Class-wise Unlearnability Analysis}
\vspace{-1mm}
We evaluate \textsc{DiffUE}’s class-specific effect by applying it solely to the ‘Frog’ class in CIFAR-10, keeping other classes untouched. A ResNet-18 model trained on this partially protected dataset shows poor classification on ‘Frog’ test samples, mispredicting them as ‘Horse’, ‘Dog’, or ‘Cat’ (Fig.~\ref{single_class}). Meanwhile, performance on untouched classes remains stable, confirming that \methodName selectively impairs learning without inducing domain shift. Extending this to three classes (‘Car’, ‘Dog’, ‘Frog’), the confusion matrix in Fig.~\ref{multi_class} shows similar degradation for protected classes, while others retain high accuracy. This demonstrates \methodName’s effectiveness under multi-class settings and its fine-grained control over unlearnability.

\begin{figure}[t]
    \centering
    \includegraphics[width=0.55\linewidth]{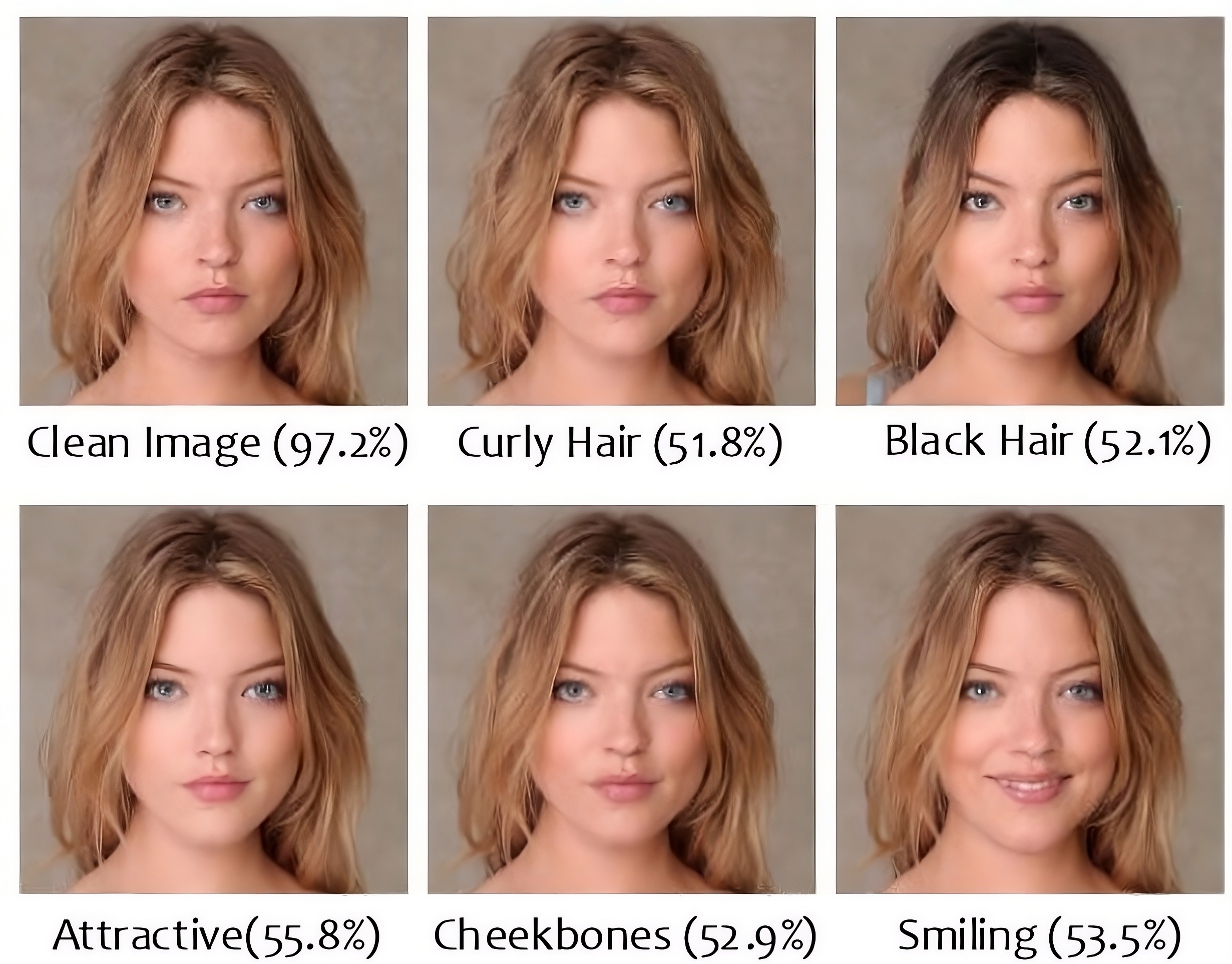}
    \vspace{-2mm}
    \caption{Defensive noise as attribute modification}
    \label{fig: Attribute_manipulation}
    \vspace{-4mm}
\end{figure}

\vspace{-2mm}
\subsection{Defensive Noise as Controlled Modifications}
\vspace{-1mm}
\methodName uses a diffusion autoencoder to guide defensive perturbations toward realistic photo edits, enhancing both unlearnability and perceptual quality. Unlike untargeted noise, which is dispersed in semantic space, control modification aligns perturbations with specific attributes for visible changes. We train a classifier on CelebA-HQ semantic codes ($z_{sem}$) to identify attributes, using its weights to steer the noise. For implementation, we use a binary gender classifier and set the perturbation radius $\rho_u$ to $0.35$, which is higher than the radius used for untargeted noise ($0.25$).

\methodName uses a diffusion autoencoder to guide defensive perturbations toward realistic photo edits, enhancing both unlearnability and perceptual quality. Unlike untargeted noise, which is dispersed in semantic space, controlled modification aligns perturbations with specific attributes for visible changes. We train a classifier on CelebA-HQ semantic codes ($z_{sem}$) to identify attributes, using its weights to steer the noise. For implementation, we use a binary gender classifier and set the perturbation radius $\rho_u$ to $0.35$, which is higher than the radius used for untargeted noise ($0.25$). The controlled semantic edits are an optional demonstration of semantic-space controllability and are not used in the default \methodName configuration.

Fig.~\ref{fig: Attribute_manipulation} illustrates the impact of attribute and style manipulations on clean test accuracy. For example, curly hair (51.8\%) and black hair (52.1\%) modifications reduce accuracy to near-random levels, demonstrating strong unlearnability. Beyond experimental validation, this method helps protect personal images from unauthorized AI training. Attribute modifications can subtly alter profile pictures without degrading quality.  By balancing privacy and image integrity, \methodName enables users to share visually appealing content while minimizing AI exploitation risks.
\begin{table}[t]
\vspace{-2mm}
\centering
\caption{Test accuracy on CIFAR-10/100 with different model architectures.}\vspace{1mm}
\resizebox{0.55\linewidth}{!}{
\begin{tabular}{llccccc}
\toprule
Dataset & Model & Clean & EM & REM & SEM & \methodName\\
\midrule
\multirow{4}{*}{CIFAR-10} & VGG-16    & 85.34 & 81.50 & 68.10 & 54.55 & \textbf{25.33}  \\
                          & RN-18     & 86.92 & 76.95 & 64.64 & 46.85 & \textbf{17.11} \\
                          & RN-50     & 87.24	& 79.80	& 71.22	& 52.00	& \textbf{20.05} \\
                          & DN-121    & 84.85	& 81.70	& 72.15	& 55.60	& \textbf{21.75} \\
\midrule
\multirow{4}{*}{CIFAR-100} & VGG-16   & 61.45 & 59.77 & 54.45 & 59.11 & \textbf{11.65}\\
                           & RN-18    & 72.19 & 64.59 & 32.22 & 26.19 & \textbf{4.92}\\
                           & RN-50    & 67.45 & 69.42 & 62.78 & 29.41 & \textbf{9.03}\\
                           & DN-121   & 52.65 & 59.89 & 66.68 & 50.70 & \textbf{10.98}\\
\bottomrule
\end{tabular}
}

\label{Tab: models}
\vspace{-6mm}
\end{table}

\vspace{-2mm}
\subsection{Transferability Analysis}
\vspace{-1mm}
To investigate the transferability of different noise methods across neural architectures, we selected ResNet-18 as the source model for generating unlearnable examples (UEs). The impact of the generated noise was evaluated on target models, including VGG-16, ResNet-50, and DenseNet-121. All transferability experiments were conducted on the CIFAR-10 and CIFAR-100 datasets, as shown in Table~\ref{Tab: models}. \methodName demonstrated superior performance in generating highly transferable UEs, achieving the lowest test accuracies in all cases. This highlights its effectiveness in surpassing existing methods such as REM and SEM.

\begin{figure}[t]
    \centering
    \includegraphics[width=0.62\linewidth]{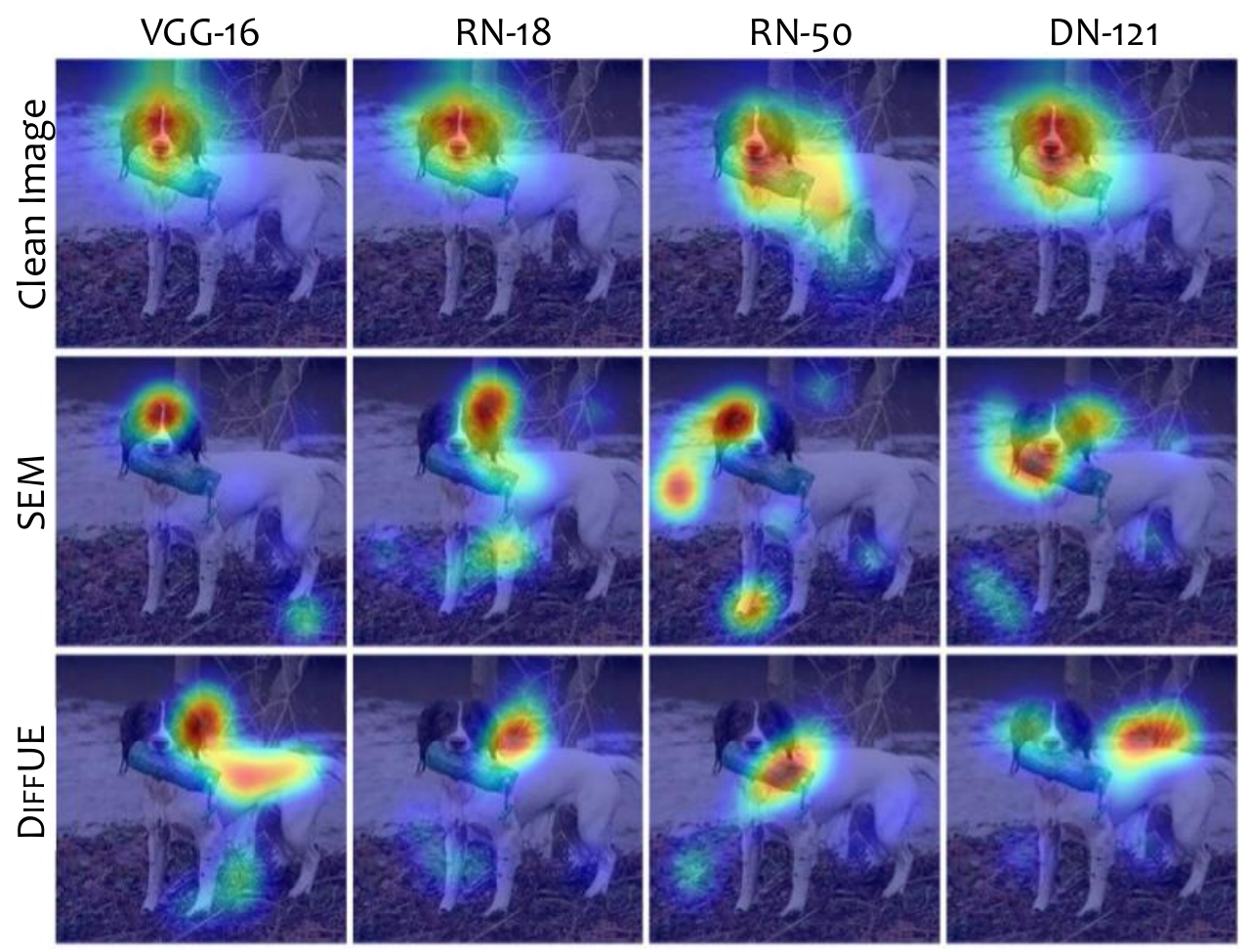}
    \caption{Comparison of the spatial regions in the input image that contribute to classification for each model when trained on clean and unlearnable examples.}
    \vspace{-2mm}
    \label{fig:gradcam}
\end{figure}

Transferring UEs to models with different architectures reduces their effectiveness due to differences in feature representations between the source and target models. As architectural disparity increases, UEs become less effective. For example, ResNet-50, which is structurally similar to ResNet-18, retains higher performance with UEs than VGG-16, which differs significantly in design. This highlights the role of architectural similarity in UE transferability. To further analyze this effect, we examine Grad-CAM visualizations of models trained on clean images and unlearnable examples. Fig.~\ref{fig:gradcam} presents results for the same input image using models trained under both settings.

When trained on clean images, all models consistently focus on the same discriminative region, the dog's face. In contrast, when trained on UEs generated using \methodName, the models shift their attention to other regions unrelated to the dog's face. Notably, even though the unlearnable examples were generated using a ResNet-18 surrogate, the altered attention patterns are consistent across all tested architectures, demonstrating the strong transferability of \methodName. By comparison, models trained on unlearnable datasets generated by SEM show less consistent attention shifts, different architectures focus on different regions, and in many cases still attend to the dog's face, indicating that SEM exhibits weaker transferability than \methodName.

\begin{table}[t]
\centering
\caption{User study comparison of \methodName with baselines across perceptual metrics.}
\resizebox{0.8\linewidth}{!}{
\begin{tabular}{lcccc}
\toprule
\textbf{Method} & \textbf{Naturalness($\uparrow$)} & \textbf{Acceptability ($\uparrow$)} & \textbf{Artifact($\downarrow$)} & \textbf{Preference($\downarrow$)} \\
\midrule
Clean   & $4.90\pm 0.10$ & $98.5\%$  &$1.10\pm0.15$ & -\\
EM      & $3.76\pm0.05$ & $56.5\%$  &$3.97\pm0.12$ & 2.86   \\
REM     & $2.91\pm0.19$ & $19.2\%$  &$4.23\pm0.07$ & 3.79   \\  
SEM     & $3.11\pm0.35$ & $61.5\%$  &$3.31\pm0.10$ & 2.13   \\
\textbf{\methodName} & $4.65\pm0.12$ & $98.5\%$  &$1.22\pm0.10$ & 1.22  \\
\bottomrule
\end{tabular}
}
\vspace{-5mm}
\label{tab:user_study}
\end{table}

\vspace{-2mm}
\subsection{User Study}
\vspace{-1mm} 
We conducted a user study to assess the perceptual quality and user preference of \methodName compared to baseline methods. A total of 56 participants were recruited via online flyers posted in photography, digital art, and content creation forums (e.g., Reddit), including 26 photographers, 13 digital artists, and 17 social media content creators. Participants independently completed the study on their own devices.

Each participant was shown $20$ randomized image sets containing portraits (3 sets contain the participant's own portraits) and landscape images, with each set containing a clean image and four unlearnable versions (EM, REM, SEM with $\rho_u=4/255$ and \methodName with $\rho_u=0.35$). They rated each image using a 5-point Likert scale on naturalness and artifact visibility, and responded to a yes/no question regarding its usability. They were also asked to rank the four UE images from each set based on overall preference. As shown in Table~\ref{tab:user_study}, \methodName achieved a naturalness score of $4.65 \pm 0.12$, closely matching the clean images ($4.90 \pm 0.10$), and maintained high usability at $98.5\%$. In contrast, baseline methods such as REM and EM received significantly lower acceptability scores of $19.2\%$ and $56.5\%$, respectively, along with higher artifact ratings. \methodName also ranked best in overall preference with an average rank of $1.22$, clearly outperforming the other baselines. These results indicate that \methodName delivers near-clean visual quality while preserving robust privacy characteristics.

\vspace{-5mm}
\section{Conclusion}
\vspace{-2mm}
To confront the challenge of developing robust unlearnable examples, we introduce \methodName, a defense framework that operates at the semantic level to preserve both unlearnability and visual realism. Built on a diffusion autoencoder, \methodName injects controlled perturbations into the semantic space, producing images that appear natural yet remain unlearnable even against advanced adversarial relearning strategies.  Experimental evaluations demonstrate that \methodName consistently achieves higher robustness than existing methods against state-of-the-art relearning attacks while preserving data usability across various applications, establishing a reliable path toward privacy-preserving data in adversarial environments.

\vspace{-5mm}
\section{Acknowledgment}
\vspace{-2mm}

We would like to thank our anonymous reviewers and shepherd for their insightful feedback. This work is supported in part by NSF CNS-2114161, ECCS-2132106, CBET-2130643, and CNS-2403529.

\bibliographystyle{splncs04}
\bibliography{main}
\end{document}